\pdfoutput=1

\documentclass[11pt]{article}

\usepackage[preprint]{acl}

\usepackage{times}
\usepackage{latexsym}
\usepackage{comment}
\usepackage[T1]{fontenc}

\usepackage[utf8]{inputenc}

\usepackage{microtype}

\usepackage{inconsolata}

\usepackage{graphicx}
\usepackage{inconsolata}
\usepackage{comment}
\usepackage{inconsolata}
\usepackage{hyperref}
\usepackage{times}
\usepackage{latexsym}
\usepackage{times}
\usepackage{latexsym}
\usepackage{graphicx}
\usepackage{lipsum}

\usepackage{float}
\usepackage{algorithm}
\usepackage{algorithmic}
\usepackage{placeins}
\usepackage{subfigure}
\usepackage{booktabs}
\usepackage{cleveref}
\usepackage{multirow}
\usepackage{xcolor, soul}
\definecolor{green}{HTML}{C2D8C8}
\definecolor{purp}{HTML}{D2C2D8}
\definecolor{red}{HTML}{F42C14}
\definecolor{dark-red}{HTML}{DE3163}
\definecolor{dark-blue}{HTML}{268EEA}
\definecolor{class}{HTML}{82204A}
\definecolor{struct}{HTML}{558C8C}
\definecolor{retrieval}{HTML}{5C85FF}
\definecolor{qa}{HTML}{EF7B45}
\usepackage{multirow}
\usepackage{rotating}
\definecolor{green}{HTML}{C2D8C8}
\definecolor{purp}{HTML}{D2C2D8}
\definecolor{red}{HTML}{F42C14}
\definecolor{dark-red}{HTML}{FFA500}
\definecolor{dark-blue}{HTML}{268EEA}
\sethlcolor{red}

\title{\textit{Best-of-L}: Cross-Lingual Reward Modeling\\ for Mathematical Reasoning }

\author{Sara Rajaee$^1$\thanks{Corresponding author: \texttt{s.rajaee@uva.nl}}, Rochelle Choenni$^2$, Ekaterina Shutova$^2$, Christof Monz$^1$ \\
  Language Technology Lab, University of Amsterdam$^1$ \\
  ILLC, University of Amsterdam$^2$
  }

\begin{document}
\maketitle
\begin{abstract}
While the reasoning abilities of large language models (LLMs) continue to advance,
 it remains unclear how such ability varies across languages in multilingual LLMs and whether different languages produce reasoning paths that complement each other. To investigate this question, we train a reward model to rank generated responses for a given question across languages. Our results show that our cross-lingual reward model substantially improves mathematical reasoning performance compared to using reward modeling within a single language, benefiting even high-resource languages. While English often exhibits the highest performance in multilingual models, we find that cross-lingual sampling particularly benefits English under low sampling budgets. Our findings reveal new opportunities to improve multilingual reasoning by leveraging the complementary strengths of diverse languages.

\end{abstract}

\section{Introduction}
\begin{figure}[!t]
    \scalebox{1}{
    \centering
\includegraphics[width=\linewidth, height=9cm]{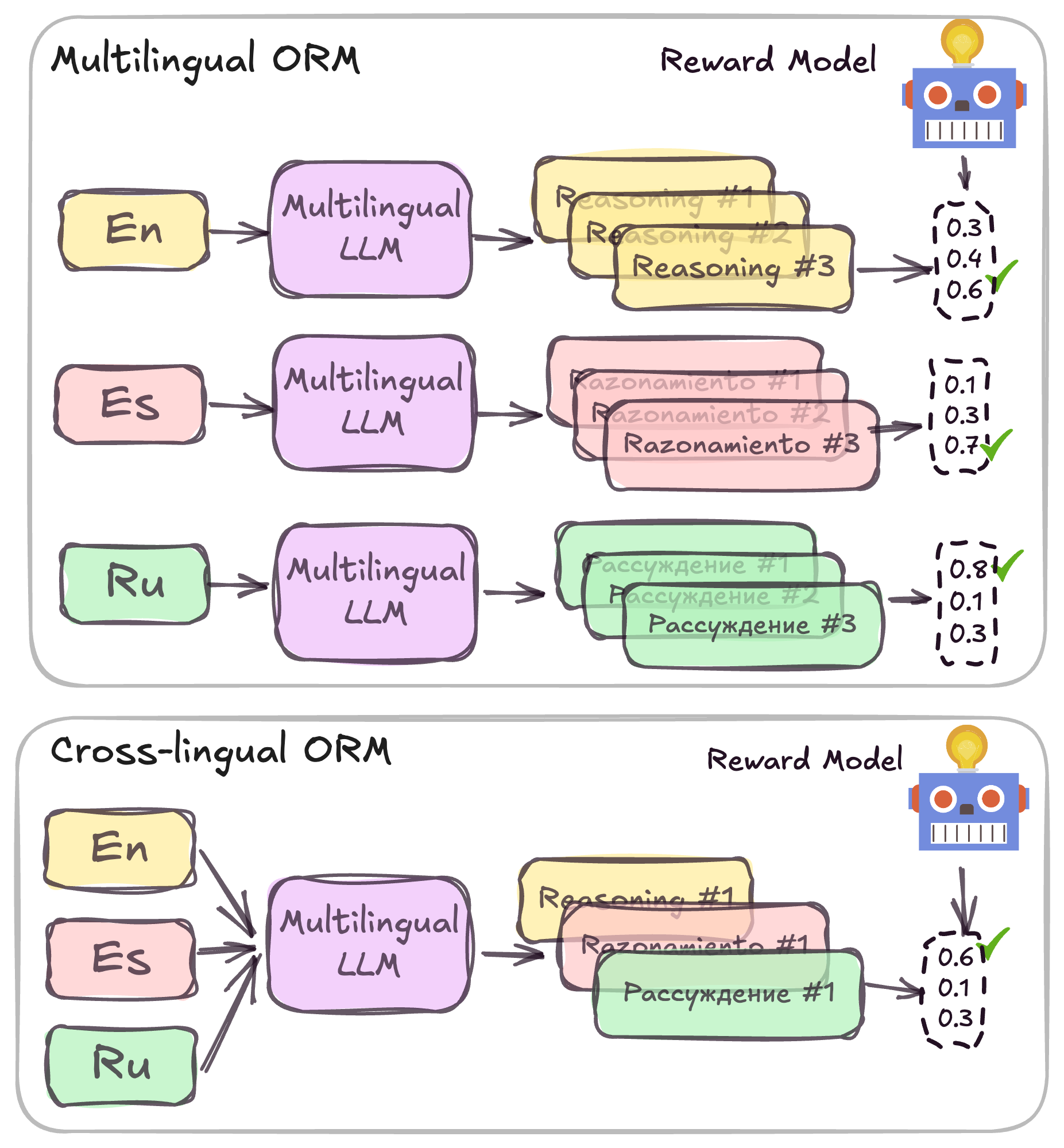}}
    \caption{An illustration of multilingual outcome reward modeling (ORM) on the top, where the verifier ranks every language's responses, and cross-lingual ORM (our framework), where the verifier ranks responses across languages for a given question.}
    \label{fig:main}
\end{figure}

Recently, many studies have focused on improving reasoning ability \cite{ranaldi-freitas-2024-self,byun-etal-2024-ares} or discovering major factors contributing to this skill \cite{ko-etal-2024-hierarchical}. Yet, reasoning research has largely centered on English models, with multilingual models receiving comparatively little attention.
Among the few, 
\citet{shi2023language} have shown that multilingual large language models (LLMs) have strong reasoning capabilities, even for underrepresented languages. Recent work has further improved the multilingual math reasoning ability of LLMs through self-consistency \cite{lai2025multidimensionalconsistencyimprovesreasoning}, multilingual instruction-tuning, \cite{chen-etal-2024-breaking,lai-nissim-2024-mcot}, and preference optimization methods \cite{she-etal-2024-mapo,dang-etal-2024-rlhf,yang2025language}.
Following the proposed studies in using reward modeling to improve the performance of math reasoning in English LLMs \cite{cobbe2021trainingverifierssolvemath,shen-etal-2021-generate-rank,hosseini2024vstar,zhang2024generative,setlur2025rewarding}, \citet{hong-etal-2025-cross} have studied the transferability of English reward models to other languages.
While \citet{wang2025demystifying} are the first to shift attention to multilingual models, they still focus on generating and scoring in-language candidate solutions, which they refer to as multilingual reward modeling.
In this paper, we instead explore the potential to generate and combine solutions in multiple different languages (see Figure~\ref{fig:main}), thereby fully exploiting the multilingual capabilities of LLMs.

To this end, we first study to what extent languages could potentially complement each other's mathematical reasoning skills. 
Interestingly, we find that even low-resource languages sometimes succeed where high-resource languages fail, suggesting that their reasoning signals could provide valuable complementary information (Figure~\ref{fig:cherry-example}). 

Motivated by the above finding, we develop a cross-lingual outcome reward modeling (ORM) framework to harness the \textit{Best-of-Languages} performance, for which we train a verifier to score multilingual reasoning across languages. 
To the best of our knowledge, we are the first to propose a cross-lingual reward model that takes advantage of complementary reasoning skills across languages. 
Our experimental results demonstrate that our framework improves performance by over 10$\%$ and 15$\%$ compared to the average performance of the naive multilingual RMs and the self-consistency baseline, respectively. Our analysis shows that increasing the number of languages improves the performance of our cross-lingual ORM.

Through an ablation study, we find that cross-language sampling even benefits English, especially under low-budget settings. Our analysis verifies that, while having English in the language pool of the cross-lingual ORM positively affects the performance, some selection of non-English pools outperforms other pools containing English, supporting our argument that languages have complementary reasoning skills in multilingual models.

\section{Methodology}

\begin{table*}[ht!]
    \scalebox{0.95}{
    \centering
    \begin{tabular}{r|cc |c | cc | cc}
        \toprule
         & En. & Avg. & SC & Multi-ORM  & Cross-ORM & Pass@8-Multi & Pass@8-Cross   \\
        \midrule
        \textbf{Aya-Expanse-8b} & 79.6 & 63.4 & 58.3 &73.3 & \textbf{83.2} & 82.4 & \textbf{93.2}\\
        \midrule
        \textbf{Llama3.1-8b} & 80.4 & 64.0 & 71.6 & 76.2  & \textbf{84.0} & 86.9 & \textbf{92.4} \\
        \midrule
        \textbf{Ministral-8b} & 82.0 & 65.1 & 70.3 & 76.4 & \textbf{87.6} & 84.3 & \textbf{93.4} \\
        \midrule
        \textbf{Qwen2.5-7b} & 85.2 & 72.2 & 74.4 & 81.3 & \textbf{92.4} & 87.2 & \textbf{96.4} \\
        \midrule
        \textbf{Phi3-7b} & 90.0 & 69.3 & 74.2 & 79.5 & \textbf{92.8} & {85.3} & \textbf{96.8}\\
        \midrule
        \textbf{Llama3.2-3b} & 72.4 & 56.0 & 63.0 & 70.3 & \textbf{77.2} & 80.3 & \textbf{88.8} \\
        \bottomrule
    \end{tabular}
    }
    \caption{Summary of the results across models and baselines. The leftmost columns represent the English performance and the average performance of all the languages. \textit{SC} denotes average self-consistency accuracy. \textit{Pass@8-cross} outperforms the average \textit{pass@8-multi}, indicating the complementary math reasoning skills across languages. Our proposed framework, \textit{Cross-ORM}, also exceeds the average \textit{Multi-ORM} accuracy by a large margin.}
    \label{tab:main_results}
\end{table*}

A popular approach in math reasoning tasks is utilizing reward models (also known as verifiers) to
evaluate the correctness of a given answer. Based
on the evaluation setup, reward models can be
process-based, where the model assesses the reasoning step by step (called PRMs)\cite{lightman2024lets,Luo2024OmegaPRM}. while outcome reward models (ORMs) evaluate the entire reasoning \cite{cobbe2021trainingverifierssolvemath,shen-etal-2021-generate-rank,hosseini2024vstar,zhang2024generative,setlur2025rewarding}. In this work, we focus
on the latter and propose a novel cross-lingual outcome reward modeling framework that leverages complementary reasoning signals across languages.

\subsection{Cross-lingual Reward Modeling}\label{sec:method}
\label{sec:verifier}
Our framework is the cross-lingual version of the \textit{Best-of-N} \cite{lightman2024lets}, ranking the generated answers for a given question across a set of languages and selecting the highest-scored one. 

Given a math question $q$ and a generated candidate answer $a$, we train a discriminative verifier to predict whether the generated reasoning is correct. More specifically, we train an LLM as the verifier using binary cross-entropy loss: $\mathcal{L}_{ORM} = - \left[ y \cdot \log(\hat{y}) + (1 - y) \cdot \log(1 - \hat{y}) \right]
$. 
At inference, we use the verifier scores to rank a set of candidate answers in different languages for a given question using the probability that the model put on the correct class, and then, we select the answer with the highest probability.


\textbf{Training set generation.} We use the GSM8K training set, including around 7.5k examples of high-quality grade school math problems created by human writers, \cite{cobbe2021trainingverifierssolvemath}, to generate our verifier training set \cite{lai-nissim-2024-mcot}. We use Google Translate version of GSM8K in 8 languages---English(en), Spanish(es), French(fr), German(de), Russian(ru), Chinese(zh), Japanese(ja), and Thai(th)---. We then prompt 3 models---the instruction-tuned version of Aya-Expanse 8B \cite{dang2024ayaexpansecombiningresearch}, Llama3.1 8B \cite{grattafiori2024llama3herdmodels}, and Qwen2.5 7B \cite{qwen2.5}---using the GSM8K training set in our 8 languages to generate responses with step-by-step reasoning. We automatically labeled the generated reasoning paths as correct or incorrect based on the correctness of the final answer. Using generations from multiple models allows us to increase the size and diversity of the training set. To make a balanced dataset, we use the same number of correct and incorrect samples for each language, resulting in a set of around 88k samples for training.

\textbf{Cross-lingual-ORM} We use the multilingual Qwen2.5-Instruct 3B model \cite{qwen2.5} as our reward model (verifier) because it has the widest officially supported language coverage among recent multilingual models. We fine-tune the verifier using the aforementioned training set for $5$ epochs, with AdamW, a learning rate of $2e$-$4$, and a batch size of $96$. Since the main task is binary classification and to make fine-tuning efficient, we fine-tune with LoRA \cite{hu2022lora} with a rank of $16$ and scaling factor of $32$. We use this as our cross-lingual outcome reward model for all experiments.

\section{Experiments}

\subsection{Experimental setups}
To study the chain-of-thought math reasoning ability of LLMs, we employ the MGSM (Multilingual Grade School Math) dataset \cite{shi2023language}, covering 11 languages, including English(en), Spanish(es), French(fr), German(de), Russian(ru), Chinese(zh), Japanese(ja), and Thai(th) with $250$ examples for each. We exclude Swahili, Telugu, and Bengali, as multilingual LLMs, including our verifier model, perform poorly on these languages. Nevertheless, our selection includes languages from diverse language families and writing scripts~\cite{lai2025multidimensionalconsistencyimprovesreasoning}. Following the original recipe of using MGSM \cite{shi2023language}, we prompt multilingual LLMs under the \textit{Native-CoT} setting using 8-shots for all experiments\footnote{We have used the evaluation harness framework for our experiments and reported exact-match scores \cite{eval-harness}.}.

\textbf{Models.}
We have carried out our analysis and experiments using a wide range of instruction-tuned multilingual models, including Aya-Expanse 8B \cite{dang2024ayaexpansecombiningresearch}, Llama3.1 8B \cite{grattafiori2024llama3herdmodels}, Qwen2.5 7B \cite{qwen2.5}, Ministral 8B\footnote{\url{https://huggingface.co/mistralai/Ministral-8B-Instruct-2410}}, phi-3 7B \cite{abdin2024phi3technicalreporthighly}, and Llama 3.2 3b\footnote{\url{https://huggingface.co/meta-llama/Llama-3.2-3B-Instruct}}.
\subsubsection{Baselines.}
We evaluate our cross-lingual ORM against the following baselines: 

\textbf{Self-consistency.} A simple, yet effective approach in chain-of-thought (CoT) prompting is self-consistency  \citep{wang2023selfconsistency,yao2023tree,kojima2022large}. This widely used baseline does the majority voting across a batch of sampled answers ($N=8$) for each language.

\textbf{Multilingual-ORM.} Also known as \textit{Best-of-N} technique, where the multilingual verifier scores $N$ different samples within a language and selects the one with the highest score \cite{wang2025demystifying}. We use $N=8$, generated with a temperature sampling of $T=0.7$, and truncated at the top-p ($p=0.95$) for all experiments (including the self-consistency baseline).


\subsection{Results and Findings}

\textbf{LLMs exhibit complementary mathematical reasoning skills across languages.}
To investigate the similarity of reasoning knowledge across languages, we employ pass@k, a well-established metric used to approximate the upper-bound performance of LLMs when generating multiple answers \cite{hosseini2024vstar,li202512surveyreasoning}. This allows us to measure the degree of potential complementarity between languages in multilingual LLMs as it considers a question solved if at least one of the answers in different languages is correct. In Table~\ref{tab:main_results}, we report pass@8 scores across languages (pass@8-Cross) and the average pass@8 scores across different samples within languages (pass@8-Multi).%
\footnote{The performance of individual languages can be found in the Appendix.} Interestingly, we observe that Pass@8-Cross outperforms the performances of individual languages, suggesting that even high-resource languages can potentially benefit from other languages.

\textbf{Sampling across languages is superior to sampling within a language.}
Building on our analysis, we employ the cross-lingual verifier described in Sec. \ref{sec:verifier} to see how languages benefit each other in practice. The middle part of Table \ref{tab:main_results} summarizes the accuracy of our cross-lingual ORM under within- and across-language settings. As shown, Cross-ORM clearly outperforms the average performance of ORM-Multi, with the largest benefits for non-English languages. These results suggest that leveraging cross-lingual signals is more effective than relying solely on monolingual reasoning, especially for underrepresented languages.

\begin{figure}[!t]
    \scalebox{1}{
    \centering
    \includegraphics[width=\linewidth, height=5.5cm]{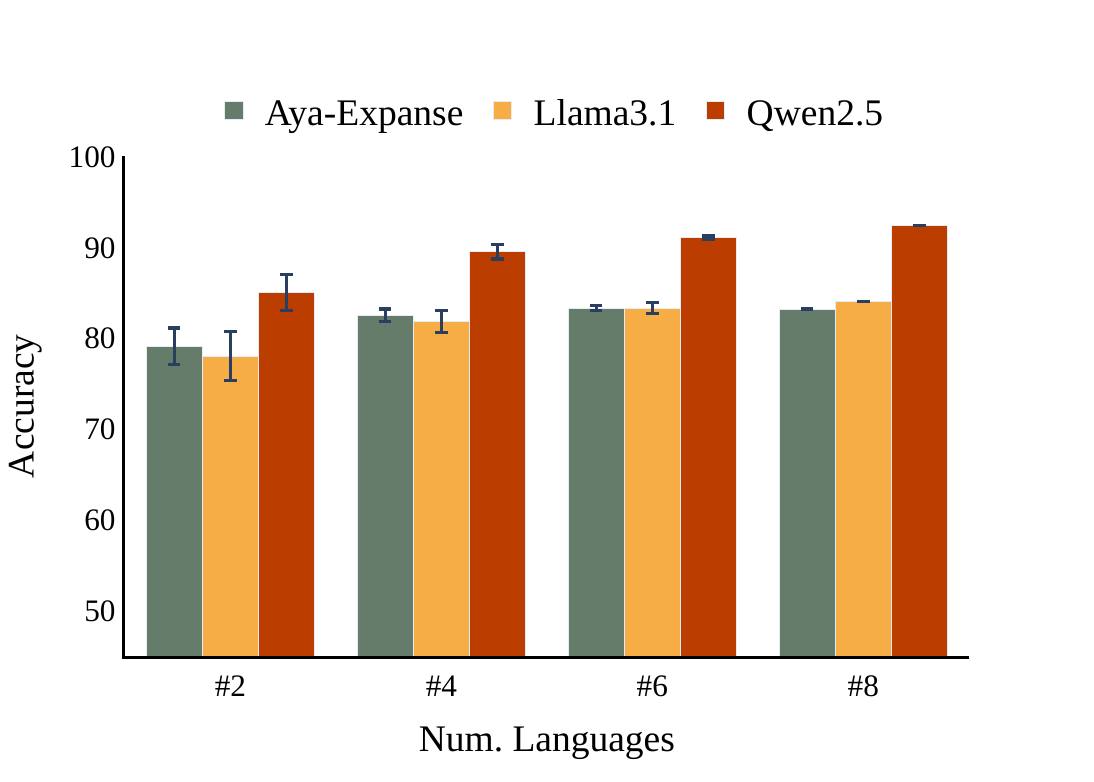}}
    \caption{The mean and standard deviation cross-lingual ORM accuracy using different numbers of languages.}
    \label{fig:language-set}
\end{figure}
\textbf{Increasing the pool of languages enhances the cross-lingual ORM performance.}
To understand the impact of language pool size, we show the average performance for all possible language combinations at different pool sizes in Figure~\ref{fig:language-set}. 
As expected, the results demonstrate that adding more languages improves cross-lingual ORM performance up to a certain point, after which the additional gains become negligible.

\begin{figure}[!t]
    \scalebox{1}{
    \centering
    \includegraphics[width=\linewidth, height=5.5cm]{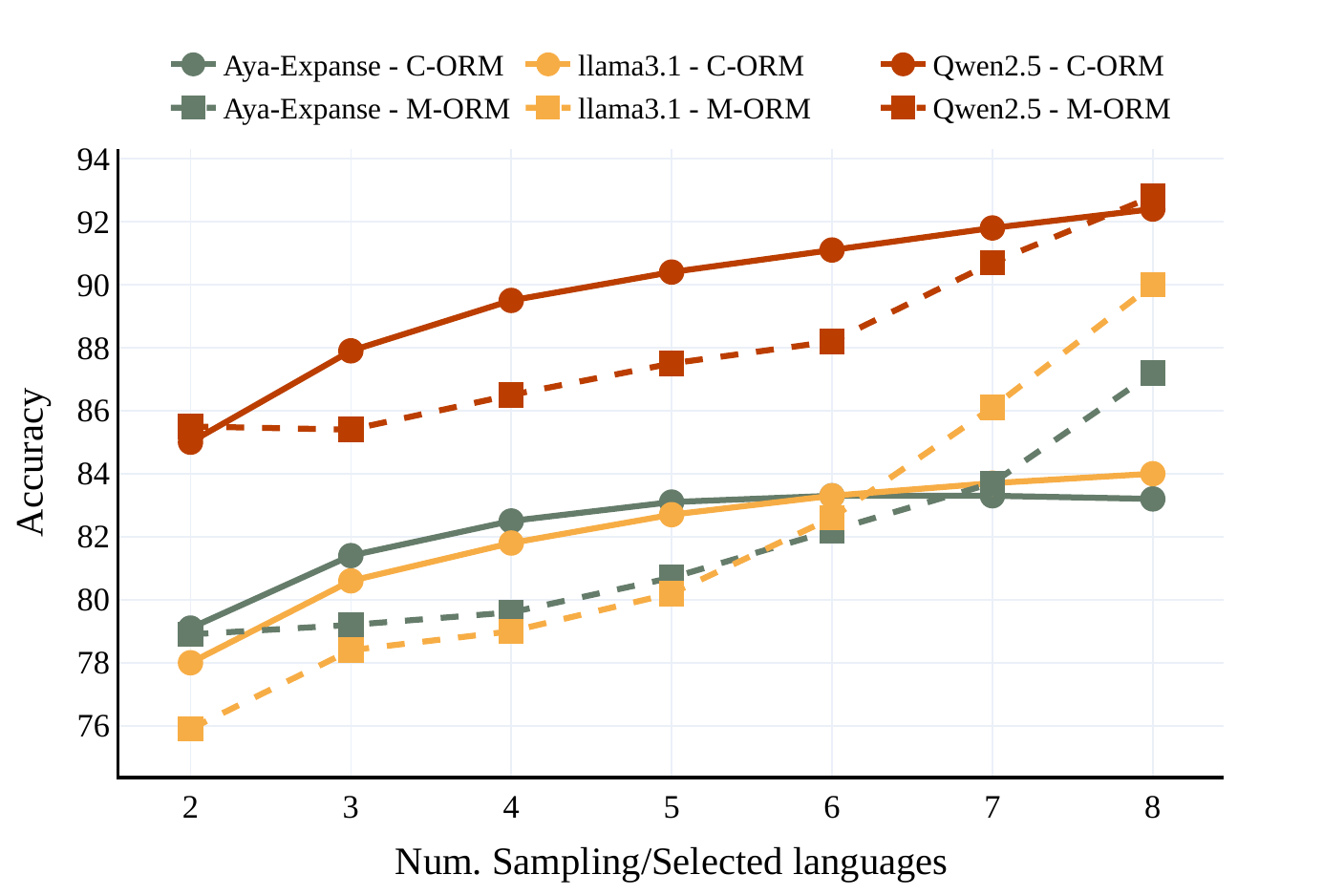}}
    \caption{Comparing the accuracy of cross-lingual ORM and multilingual ORM of English using the same number of languages and samples.}
    \label{fig:sampling-language}
\end{figure}

\textbf{Sampling across languages benefits English as well.}
While our earlier analysis shows that cross-lingual ORM exceeds the average performance of multilingual ORM, its accuracy still lags behind that of English ORM.
To better understand under what conditions other languages might benefit English, we compare the performance of English ORM (i.e., generating multiple answers in English) and cross-lingual ORM under different sampling budgets in Figure~\ref{fig:sampling-language}. 
Based on the results, we observe that cross-lingual ORM outperforms English ORM at low sampling budgets. 
However, this advantage fades as the number of samples increases. 
We suspect that additional sampling from other languages becomes redundant once English samples already cover a wide range of reasoning paths.

\begin{figure}[!t]
    \scalebox{1}{
    \centering
    \includegraphics[width=\linewidth, height=5.5cm]{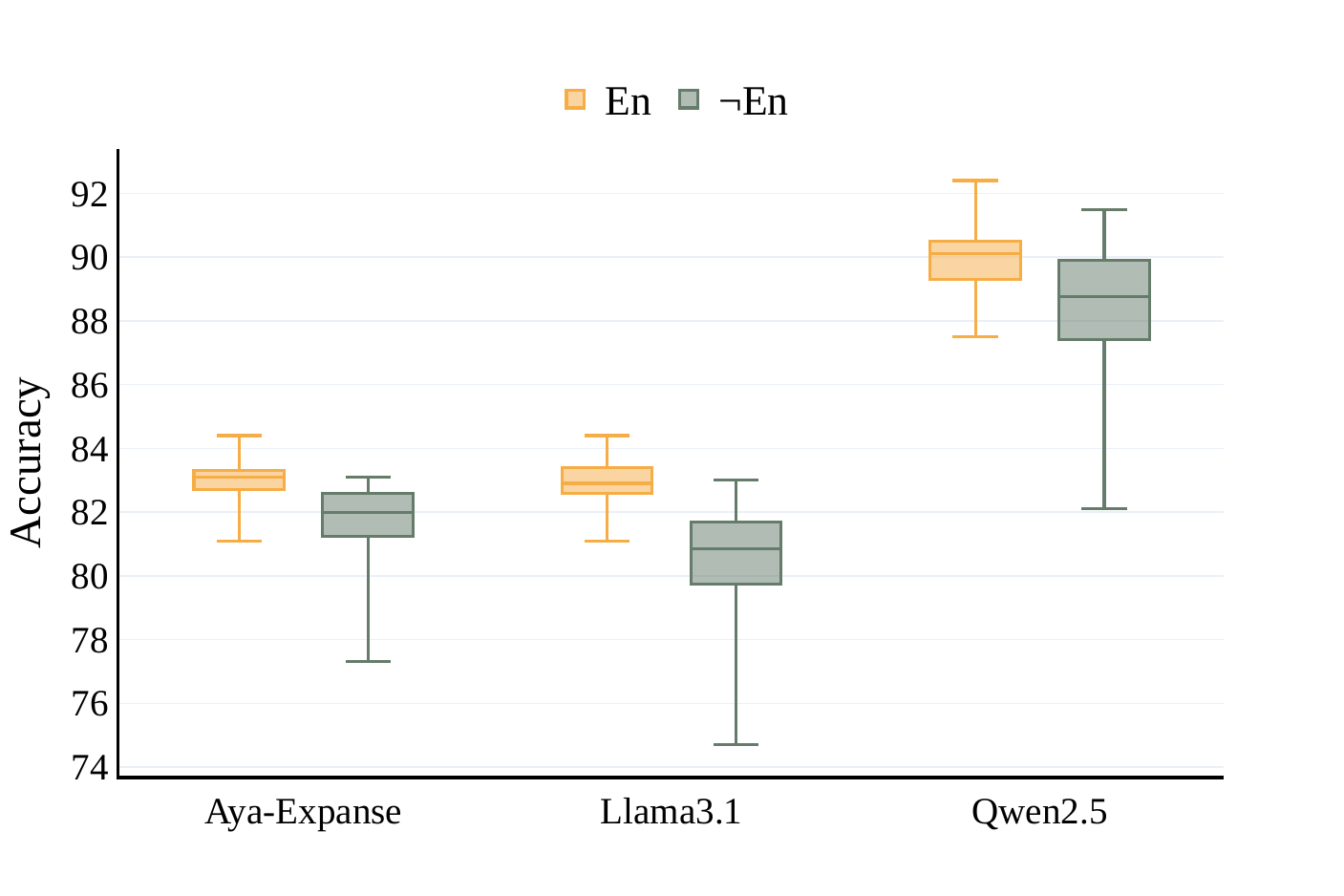}}
    \caption{Average cross-lingual ORM performance across language pools of size 2–7, with and without English. English generally helps, but some non-English sets outperform English-inclusive ones.}
    \label{fig:en-role}
\end{figure}

\textbf{Including English in language pools is generally helpful, yet it does not always lead to superior performance.}
To examine the effect of including English in the language pools for the cross-lingual ORM setup, we report the mean and standard deviation of accuracy across all possible language pools with a size of 2 to 7 with and without English in Figure~\ref{fig:en-role}. 
As expected, including English generally improves cross-lingual ORM performance. 
However, this is not always the case; some language pools without English perform better than certain groups that include English, as reflected in the standard deviation of the non-English groups.


\section{Conclusion}
In this paper, we present a novel cross-lingual reward modeling framework that effectively leverages complementary mathematical reasoning skills across languages in multilingual LLMs. Our experiments show that cross-lingual reward modeling benefits even high-resource languages like English under low-budget inference settings. Furthermore, our findings show that 
languages mutually enhance each other’s reasoning abilities. Our results pave the way for future research into the similarities and differences of reasoning patterns across languages to improve multilingual reasoning models. 

\section{Limitations}
A limitation of our work is that we focused solely on math reasoning tasks, and future research could explore other downstream tasks to broaden the applicability of our approach. Additionally, we used only eight languages, so expanding the number and diversity of languages would be important to further enhance our understanding of multilingual reasoning in LLMs. Another limitation is that we did not investigate the underlying reasoning patterns across languages, which could provide valuable insights for improving multilingual reasoning performance.

\bibliography{anthology,custom}
\appendix
\section{Full Results}
\begin{figure*}[ht]
    \scalebox{1}{
    \centering
\includegraphics[width=\linewidth, height=12cm]{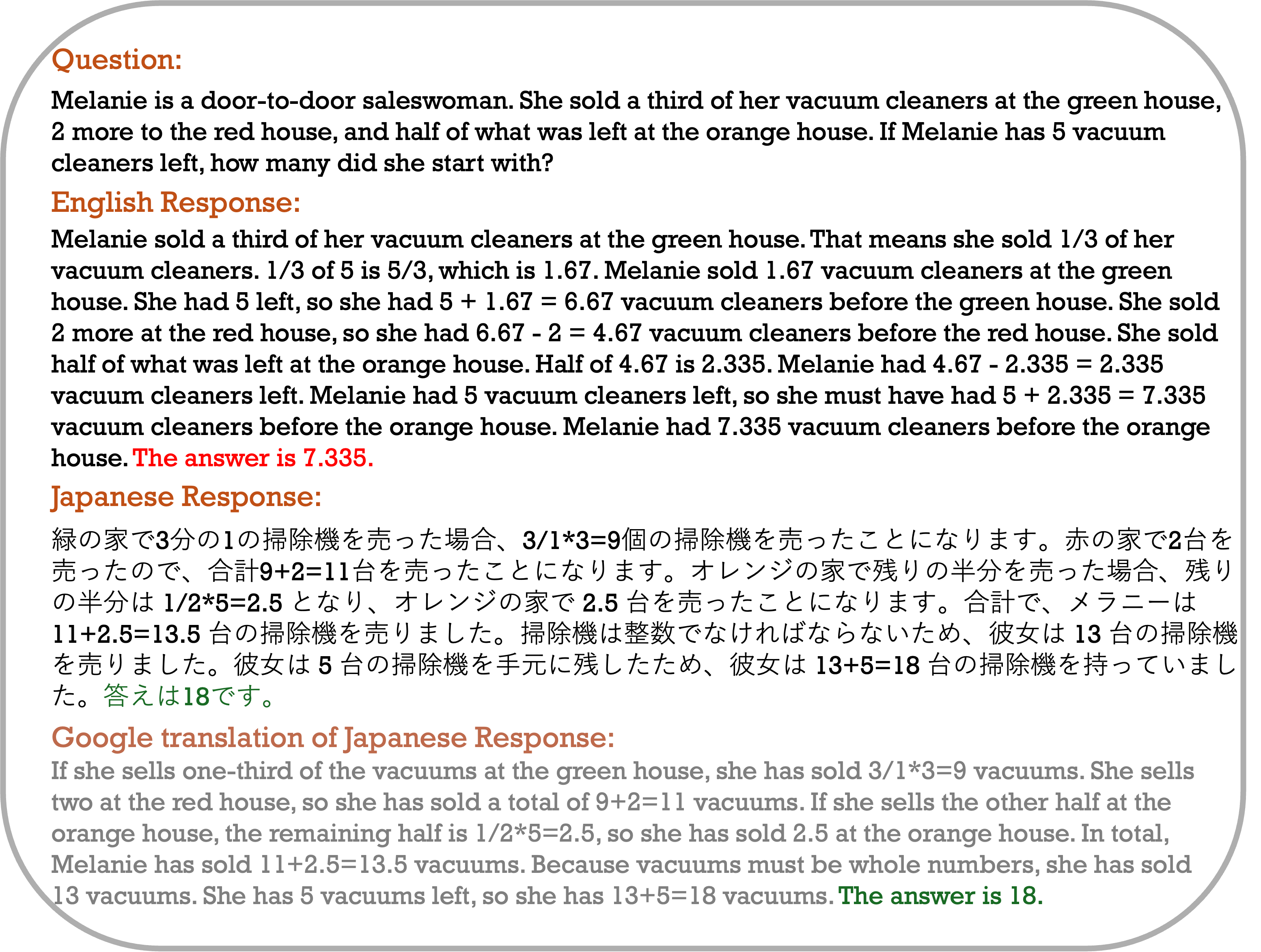}}
    \caption{A cherry-picked example of Llama 3.1’s responses to the same question in English and Japanese, where the English answer is incorrect but the Japanese response is correct, illustrating the complementary reasoning skills across languages.}
    \label{fig:cherry-example}
\end{figure*}

\begin{table*}[ht!]
    \centering
    \scalebox{0.8}{
    \begin{tabular}{r|llllllll|c}
    \toprule
    & en & fr & es & de & ru & zh & ja & th & avg. \\
    \midrule
        \multicolumn{10}{c}{\textbf{Aya-Expanse-8b}} \\

    \midrule
    CoT & 76.6 {\scriptsize$\pm$1.1} & 65.1 {\scriptsize$\pm$2.0} & 73.0 {\scriptsize$\pm$1.8} & 68.8 {\scriptsize$\pm$1.9} & 67.6 {\scriptsize$\pm$2.5} & 63.7 {\scriptsize$\pm$0.8} & 57.7 {\scriptsize$\pm$1.2} & 19.7 {\scriptsize$\pm$1.8} & 61.5 \\
    SC & 84.0 & 70.8 & 76.8 & 76.8 & 73.6 & 71.2 & 67.2 & 23.2 & 58.3 \\
    \midrule
        \multicolumn{10}{c}{\textbf{Llama3.1-8b}} \\
    \midrule
    CoT & 75.0\scriptsize$\pm$1.7 & 60.4\scriptsize$\pm$1.4 & 66.8\scriptsize$\pm$1.9 & 59.6\scriptsize$\pm$3.8 & 61.1\scriptsize$\pm$2.8 & 57.5\scriptsize$\pm$2.3 & 46.9\scriptsize$\pm$2.6 & 47.4\scriptsize$\pm$1.1 & {59.4} \\
    SC & 84.8 & 72.4 & 80.0 & 74.0 & 74.0 & 69.6 & 58.8 & 59.2 & 71.6 \\
    \midrule
        \multicolumn{10}{c}{\textbf{Ministral-8b}} \\
    \midrule
    CoT & 77.4\scriptsize$\pm$1.5 & 65.1\scriptsize$\pm$0.7 & 71.4\scriptsize$\pm$1.3 & 64.9\scriptsize$\pm$1.8 & 65.4\scriptsize$\pm$0.8 & 58.8\scriptsize$\pm$1.7 & 43.8\scriptsize$\pm$1.9 & 44.5\scriptsize$\pm$1.1 & {61.4} \\
    SC & 87.2 & 72.0 & 80.0 & 74.0 & 75.2 & 67.6 & 55.2 & 54.0 & 70.3 \\
    \midrule
        \multicolumn{10}{c}{\textbf{Qwen2.5-7b}} \\
    \midrule
    CoT & 84.3\scriptsize$\pm$1.0 & 70.2\scriptsize$\pm$1.8 & 76.3\scriptsize$\pm$2.3 & 63.8\scriptsize$\pm$1.2 & 69.0\scriptsize$\pm$1.7 & 67.8\scriptsize$\pm$1.7 & 63.7\scriptsize$\pm$1.4 & 51.8\scriptsize$\pm$1.3 & {68.4} \\
    SC & 89.2 & 73.6 & 82.8 & 70.8 & 73.6 & 75.6 & 70.0 & 59.6 & 74.4 \\
    \midrule
        \multicolumn{10}{c}{\textbf{Phi3-7b}} \\
    \midrule
    CoT & 87.6\scriptsize$\pm$1.8 & 77.0\scriptsize$\pm$2.3 & 83.7\scriptsize$\pm$1.7 & 77.3\scriptsize$\pm$0.9 & 71.6\scriptsize$\pm$1.8 & 69.2\scriptsize$\pm$2.3 & 58.0\scriptsize$\pm$1.4 & 18.2\scriptsize$\pm$1.3 & {67.8} \\
    SC & 92.8 & 84.4 & 88.4 & 83.6 & 82.8 & 74.4 & 65.6 & 23.6 & 74.2 \\
    \midrule
        \multicolumn{10}{c}{\textbf{Llama3.2-3b}} \\
    \midrule
    CoT & 65.6\scriptsize$\pm$1.1 & 51.0\scriptsize$\pm$2.7 & 56.4\scriptsize$\pm$1.0 & 52.4\scriptsize$\pm$1.1 & 53.1\scriptsize$\pm$2.4 & 48.8\scriptsize$\pm$2.1 & 32.0\scriptsize$\pm$1.7 & 44.1\scriptsize$\pm$1.9 & {50.4} \\
    SC & 79.2 & 60.4 & 70.8 & 67.2 & 65.6 & 62.8 & 44.4 & 55.6 & 63.0 \\
    \bottomrule
    \end{tabular}}
    \caption{Vanilla Chain-of-thought(CoT) performance and self-consistency (SC) on MGSM. }
\end{table*}

\begin{table*}[ht]
\centering
\begin{tabular}{r|cccccccc|c}
\toprule
 & {en} & {fr} & {es} & {de} & {ru} & {zh} & {ja} & {th} & {avg.} \\
\midrule
Aya-Expanse-8b  & 94.4 & 85.6 & 90.8 & 89.2 & 87.6 & 84.4 & 82.4 & 44.4 & {82.3} \\
Llama3.1-8b  & 94.8 & 85.2 & 92.4 & 89.2 & 92.0 & 86.8 & 74.8 & 80.0 & {86.9} \\
Ministral-8b & 94.4 & 85.6 & 91.6 & 85.2 & 86.4 & 86.4 & 73.2 & 71.6 & 84.3 \\
Qwen2.5-7b & 95.6 &  84.0 & 94.4 & 85.2 & 87.6 & 91.2 & 84.0 & 75.6 & {81.3} \\
phi3-3b & 96.4 & 92.0 & 95.6 & 90.8 & 92.0 & 88.4 & 84.8 & 42.0 & 85.3 \\
Llama3.2-3b & 91.6 & 78.8 & 85.6 & 83.2 & 82.8 & 80.8 & 66.0 & 73.6 & 80.3 \\
\bottomrule
\end{tabular}
\caption{Comparison of Pass@8-Multi across different languages on the MGSM task.}
\end{table*}

\begin{table*}[ht]
\centering
\begin{tabular}{r|cccccccc|c}
\toprule
 & {en} & {fr} & {es} & {de} & {ru} & {zh} & {ja} & {th} & {avg.} \\
\midrule
Aya-Expanse-8b  & 87.2 & 76.4 & 83.2 & 76.8 & 78.4 & 76.0 & 71.2 & 37.2 & {73.3} \\
Llama3.1-8b  & 90.0 & 77.6 & 84.0 & 60.4 & 80.0 & 78.8 & 68.4 & 70.0 & {76.2} \\
Ministral-8b & 91.2 & 76.0 & 87.6 & 77.6 & 77.2 & 78.0 & 60.4 & 62.8 & 76.4 \\
Qwen2.5-7b & 92.8 & 80.0 & 86.4 & 74.8 & 82.4 & 85.2 & 76.0 & 72.4 & {81.3} \\
Phi3-7b & 94.0 & 86.8 & 91.2 & 86.0 & 84.0 & 83.2 & 75.6 & 35.2 & 79.5 \\
Llama3.2-3b & 83.2 & 70.0 & 76.4 & 62.4 & 75.2 & 70.8 & 56.8 & 67.2 & 70.3 \\

\bottomrule
\end{tabular}
\caption{Comparison of Multi-ORM across different languages on the MGSM task.}
\end{table*}

\begin{figure}[h]
    \scalebox{1}{
    \centering
    \includegraphics[width=\linewidth, height=5.5cm]{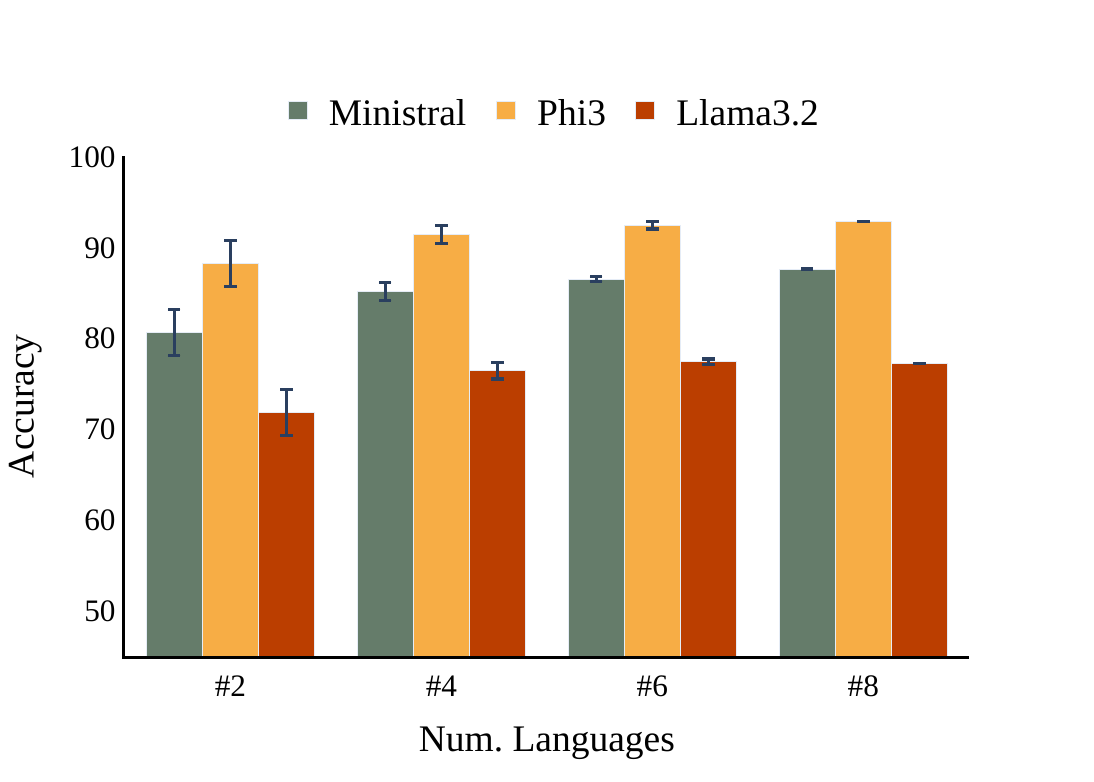}}
    \caption{The average and standard deviation cross-lingual ORM performance using different numbers of languages.}
    \label{fig:language-set2}
\end{figure}

\begin{figure}[h]
    \scalebox{1}{
    \centering
    \includegraphics[width=\linewidth, height=5.5cm]{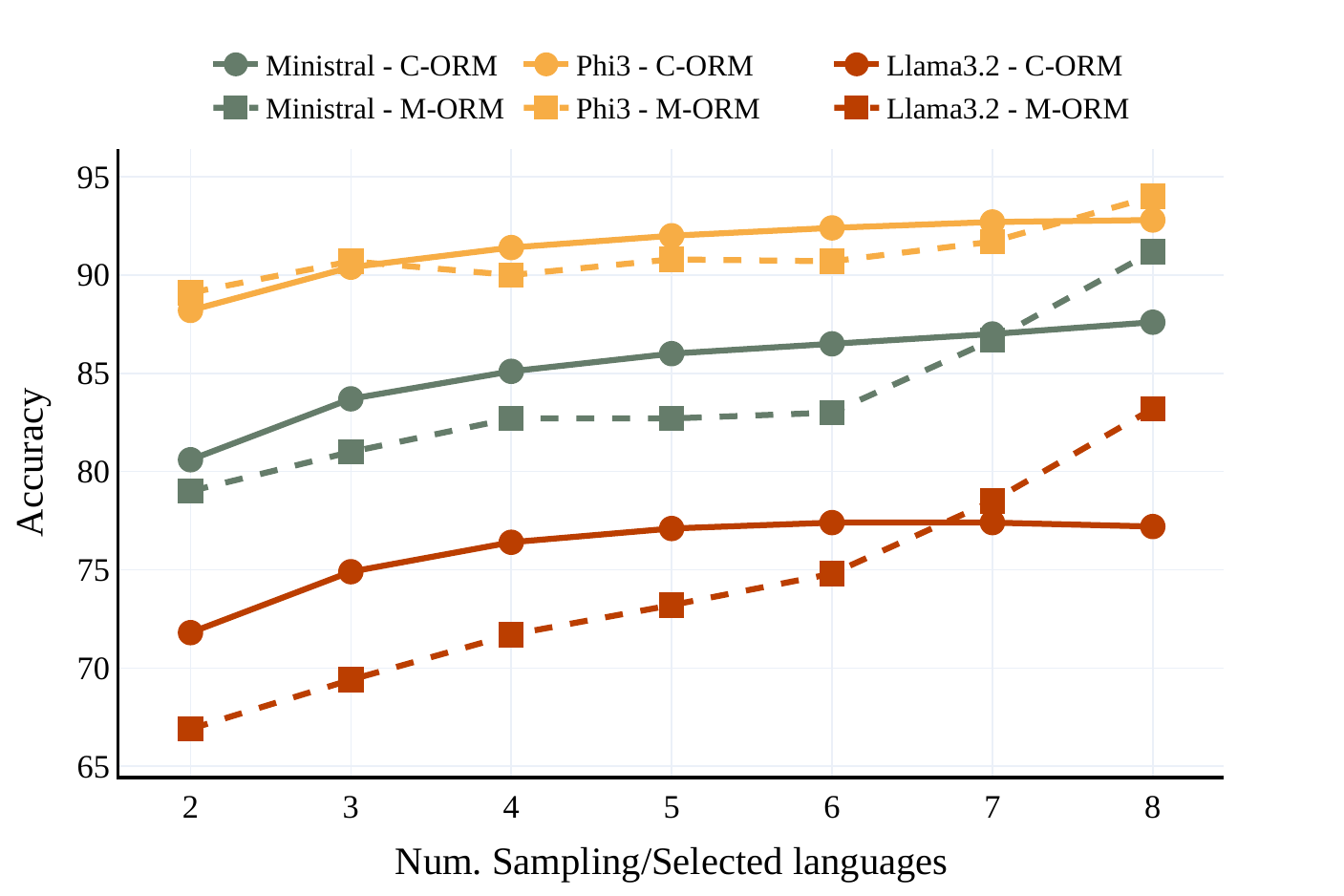}}
    \caption{Comparing the accuracy of cross-lingual ORM and multilingual ORM of English using the same number of languages and samples.}
    \label{fig:sampling-language2}
\end{figure}
\begin{figure}[h!]
    \scalebox{1}{
    \centering
    \includegraphics[width=\linewidth, height=5.5cm]{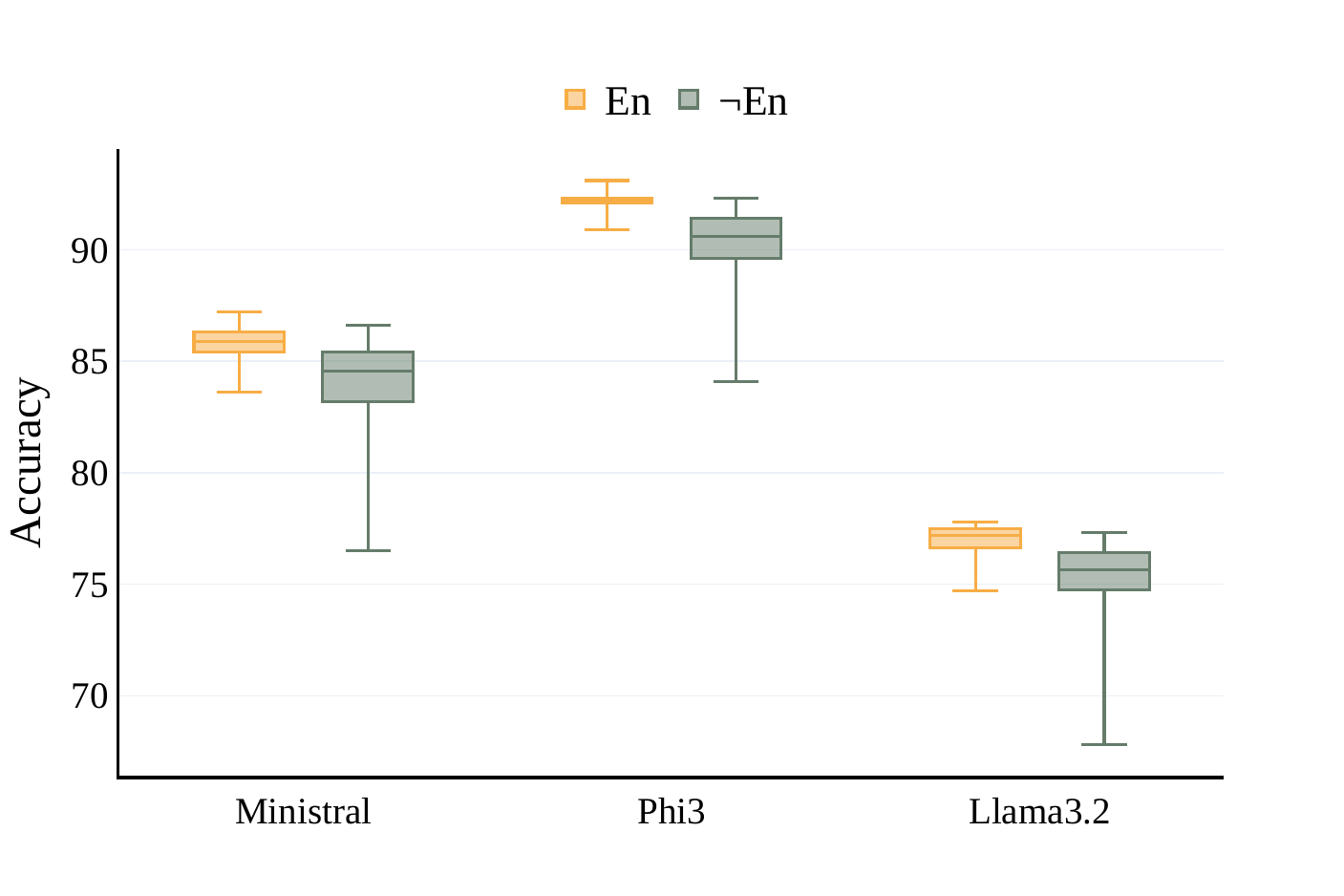}}
    \caption{Comparing the role of English on the performance.}
    \label{fig:en-role2}
\end{figure}

\end{document}